\pdfoutput=1

\documentclass[11pt]{article}

\usepackage[final]{acl}

\usepackage{times}
\usepackage{latexsym}

\usepackage[T1]{fontenc}

\usepackage[utf8]{inputenc}

\usepackage{microtype}

\usepackage{amsmath,amsfonts,bm}

\def\eqref#1{equation~\ref{#1}}

\def\1{\bm{1}}

\DeclareMathAlphabet{\mathsfit}{\encodingdefault}{\sfdefault}{m}{sl}
\SetMathAlphabet{\mathsfit}{bold}{\encodingdefault}{\sfdefault}{bx}{n}

\definecolor{redlinkcolor}{rgb}{0.79607843, 0.25098039, 0.25882353}
\definecolor{bluecitecolor}{rgb}{0,0.36,0.69}
\usepackage{hyperref}  
\usepackage{enumitem}
\usepackage{graphicx} 
\usepackage{amssymb}
\usepackage{dsfont}
\usepackage{amsmath}
\usepackage{algorithm}
\usepackage{algorithmicx}
\usepackage{algpseudocode}
\usepackage{amsfonts}
\usepackage{graphicx}
\usepackage{booktabs}
\usepackage{colortbl}
\usepackage{multirow}
\usepackage{xcolor}
\definecolor{lightblue}{RGB}{243,251,255}
\definecolor{lightyellow}{RGB}{255,252,244}
\usepackage{wrapfig}
\usepackage{nicematrix}

\newcommand{\ours}{\textsc{CoWest}}
\usepackage{inconsolata}

\usepackage{graphicx}

\title{Synergistic Weak-Strong Collaboration by Aligning Preferences}

\author{Yizhu Jiao$^{1,2}$, Xuchao Zhang$^{1}$, Zhaoyang Wang$^{3}$, Yubo Ma$^{4}$, Zhun Deng$^{3}$, Rujia Wang$^{1}$, \\ \textbf{Chetan Bansal$^{1}$, Saravan Rajmohan$^{1}$, Jiawei Han$^{1}$, Huaxiu Yao$^{1,3}$} \\
  $^{1}$Microsoft Research, $^{2}$University of Illinois Urbana-Champaign, \\ 
  $^{3}$University of North Carolina at Chapel Hill, $^{4}$Nanyang Technological University \\
  \texttt{yizhuj2@illinois.edu, xuchaozhang@microsoft.com, huaxiu@cs.unc.edu} }

\begin{document}
\maketitle
\begin{abstract}
Current Large Language Models (LLMs) excel in general reasoning yet struggle with specialized tasks requiring proprietary or domain-specific knowledge. Fine-tuning large models for every niche application is often infeasible due to black-box constraints and high computational overhead. To address this, we propose a collaborative framework that pairs a specialized weak model with a general strong model. The weak model, tailored to specific domains, produces initial drafts and background information, while the strong model leverages its advanced reasoning to refine these drafts, extending LLMs' capabilities to critical yet specialized tasks. To optimize this collaboration, we introduce a collaborative feedback to fine-tunes the weak model, which quantifies the influence of the weak model's contributions in the collaboration procedure and establishes preference pairs to guide preference tuning of the weak model. We validate our framework through experiments on three domains. We find that the collaboration significantly outperforms each model alone by leveraging complementary strengths. Moreover, aligning the weak model with the collaborative preference further enhances overall performance. \href{https://github.com/yzjiao/CoWest}{The code is publicly available}.

\end{abstract}

\section{Introduction}
The rapid evolution of Large Language Models (LLMs) \citep{zhao2023survey, chang2024survey} has exhibited remarkable proficiency in general reasoning \citep{kojima2022large, zheng2023judging}, problem-solving \citep{lewkowycz2022solving, yao2024tree}, and natural language understanding \citep{wei2022emergent}. These models have demonstrated the ability to perform a broad range of tasks across diverse domains, often with minimal task-specific training. However, their immense size and general-purpose training can make them less effective in specialized tasks or domains that are underrepresented in their training data or require access to proprietary information \citep{fu2023specializing}. This limitation poses a significant challenge: how can we extend the problem-solving spectrum of LLMs to encompass these niche but critical tasks? 

Directly training or fine-tuning large models for every specific domain or task is often impractical due to the following two key reasons. First, some popular LLMs (e.g., GPT-4 \citep{achiam2023gpt}, Gemini \citep{team2023gemini}) are black-box models, with their internal parameters inaccessible for modification. Even when fine-tuning is possible, it can be costly and raises concerns about scalability as models continue to grow in size, such as those models exceeding 70 billion parameters.  
Additionally, fine-tuning LLMs on private data can pose security and privacy risks. 
Specifically, fine-tuning requires exposing the model to potentially sensitive data, which could inadvertently be memorized or leaked through the model's outputs. This exposure creates a risk of violating data privacy regulations and necessitates robust measures to ensure data confidentiality and compliance.

\begin{figure}[!tbp]
    \centering    
    \includegraphics[width=0.5\textwidth]{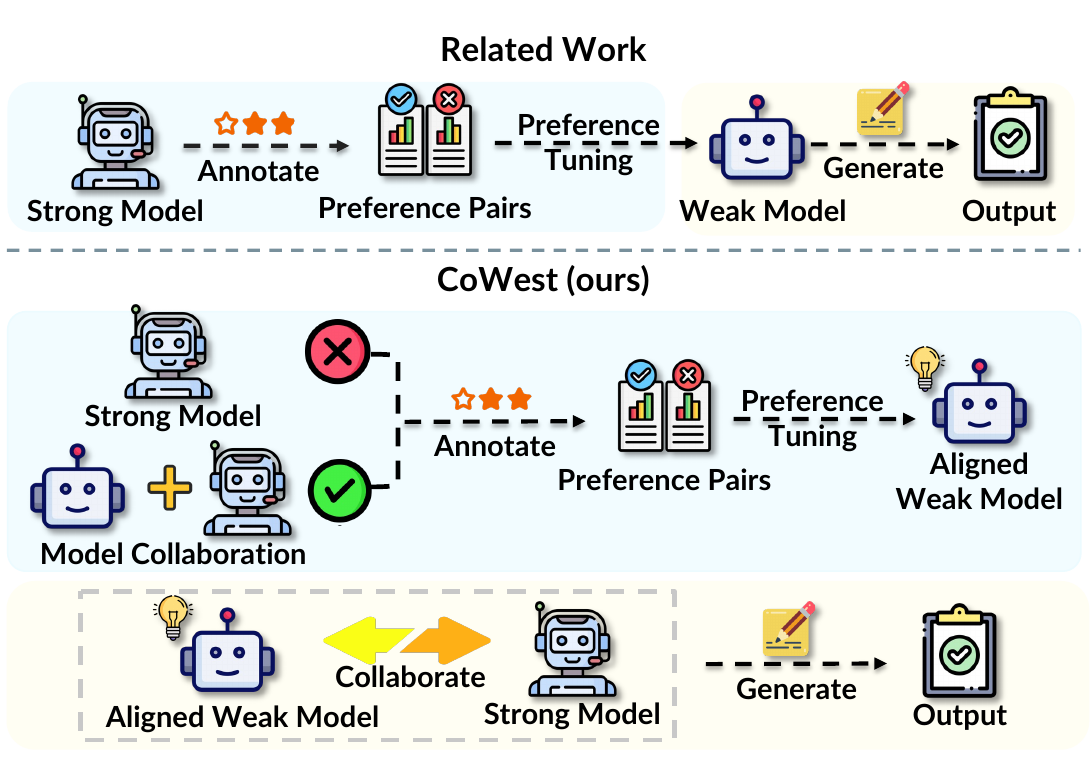}
    \caption{Comparison of our method and the related work during \colorbox{lightblue}{training} and \colorbox{lightyellow}{inference}.}
    \label{fig:intro}
\end{figure}

To overcome these challenges, we aim to leverage a collaborative framework that synergizes a small-sized weak model with a large-sized strong model. In this paradigm, the weak model is tailored with specialized problem-solving abilities in specific domains. Conversely, the strong model boasts robust general capabilities, excelling in tasks that require broad knowledge and advanced reasoning. By orchestrating a collaboration between these two models, we leverage their complementary strengths to tackle specific tasks more effectively than  
either could achieve independently. The weak model contributes domain-specific insights and preliminary solution drafts, while the strong model refines these drafts using its advanced reasoning capabilities.

While a few existing works have explored forms of weak and strong model collaboration~\citep{ juneja-etal-2023-small, shen2024learning}, they often predefine the interaction mechanisms—for example, the strong model directly receives knowledge pieces in a predefined form generated by the weak model~\citep{juneja-etal-2023-small}. However, the most effective interaction strategy can vary depending on the specific scenario or models involved. Moreover, prior approaches typically focus on individual feedback, where the single model provides the feedback to finetune another model. They overlook the potential benefits of feedback from the collaboration procedure (as shown in Figure \ref{fig:intro}), which helps the weak model understand the strong model's preferences and to enhance the mutual cooperation between the two models.

In this paper, we introduce an innovative framework for dynamic weak-strong model collaboration. Our approach harnesses the specialized knowledge of a knowledge-intensive weak model to generate detailed initial drafts and background information. The strong model then applies its robust general reasoning capabilities to refine these drafts, effectively merging the strengths of both models. To further optimize this collaborative interaction, we implement a feedback loop, which fine-tunes the weak model based on the strong model’s preferences, creating an adaptive and synergistic interaction that continuously improves. We evaluate the impact of the weak model's contributions on collaborative performance and construct preference pairs for preference tuning the weak model. This data-driven strategy allows us to amplify beneficial contributions from the weak model and minimize detrimental ones, thereby fostering a mutually beneficial interaction.

We validate our framework through experiments on three datasets, yielding several key findings:  (1) Significant Performance Gains through Collaboration: The collaboration between the weak and strong models significantly outperforms each model operating independently, demonstrating the effectiveness of leveraging complementary strengths. 
(2) Effectiveness of Finetuning Weak Model with Strong Counterpart Preference for Mutually Beneficial Interaction: Incorporating feedback from the strong model to fine-tune the weak model enhances the overall effectiveness of the collaboration. This iterative refinement allows the weak model to align closely with the strong model's preferences and reasoning patterns.
(3) Enhanced Gains with Strong Models of High General Capability: The collaborative gains are  substantial when the strong model possesses sufficient general abilities. Merely having a strong model that is better than the weak model does not guarantee mutual improvement; the strong model's capacity to correct the weak model's outputs is critical.

\section{Related Work}

\subsection{Enhancing LLMs for Solving Specialized Problems}

Addressing specialized problems beyond the generalist training of LLMs has been a key research focus. A common approach is retrieval-augmented generation, where an LLM queries external sources for domain-specific information to enhance responses \citep{DBLP:conf/icml/GuuLTPC20, DBLP:journals/corr/abs-2208-03299, DBLP:conf/iclr/Sun0TYZ23, DBLP:conf/emnlp/JiangXGSLDYCN23, DBLP:conf/acl/ZhangFC24}.  
However, these methods often provide static context for LLM to generate responses without further refinement or learning from that context. This static nature can lead to less adaptability in complex, evolving problem-solving scenarios.
Another approach uses small models for domain-specific processing to guide LLMs. This includes weak-to-strong generalization, where a strong model learns from the supervision of a weaker one \citep{DBLP:conf/icml/BurnsIKBGACEJLS24, DBLP:journals/corr/abs-2405-15116, DBLP:journals/corr/abs-2407-13647, guo2024improving, zheng2024weak, sun2024easy}. However, this often requires access to the strong model’s parameters, which can be a challenge for black-box systems. Other methods prompt LLMs with the outputs of small models to enhance performance on niche tasks \citep{DBLP:conf/acl/XuXWLZM24, liu2024panda}. Additionally, small models can act as intermediaries by identifying relevant context or splitting problems into subtasks, thereby reducing complexity for the larger model \citep{juneja-etal-2023-small, shen2024learning}.
While these methods improve LLM performance on specialized tasks, they often rely on static interaction schemes, limiting the weaker model to retrieval or prompting. In contrast, we introduce a dynamic feedback loop between weak and strong models, fostering adaptive collaboration that evolves with the task.

\subsection{Multi-Model Collaboration}

Model collaboration explores the effective utilization of the collaborative strengths of multiple LLMs
These works are generally classified into three categories: Merging, Ensemble and Alignment \citep{lu2024merge}. 
Model merging combines the parameters of various LLMs into a cohesive model, requiring compatibility of parameters within a linear framework \citep{DBLP:conf/icnn/SzymanskiL93, DBLP:journals/jmlr/FedusZS22, DBLP:journals/corr/abs-2401-04088, DBLP:journals/corr/abs-2309-09117, DBLP:conf/emnlp/DengR23, ji2024aligner}. 
Moreover, model ensemble leverages the outputs of different LLMs to produce unified outcomes, focusing less on the parameters of the individual models \citep{DBLP:journals/corr/abs-2309-15789, DBLP:conf/acl/Jiang0L23, srivatsa2024harnessing}. But previous research typically concentrated on interactions between models of comparable size or employed a fixed interaction mechanism for poor task adaptation. 
Furthermore, alignment can be viewed as a specialized form of model collaboration, where a specialized reward model provides feedback to guide a target model toward desired objectives \citep{tao2024your, DBLP:conf/icml/0001PMMFLBHCRP24}. 
However, these approaches typically focus on individual feedback and overlook the potential benefits of feedback from the collaboration procedure, which can enhance mutual cooperation between the two models.

\section{The Proposed Method - \ours}

We introduce \ours, a \underline{Co}llaboration method between \underline{We}ak and \underline{St}rong models that harnesses their complementary strengths and align the weak model with collaborative feedback to improve collaboration performance. 
Specifically, during inference, our approach harnesses the specialized knowledge of a knowledge-intensive weak model to generate detailed initial drafts and background information. The strong model then applies its robust general reasoning capabilities to refine these drafts, effectively merging the strengths of both models. To further optimize this collaborative interaction, during training, we implement a feedback loop, which fine-tunes the weak model based on the strong model’s preference, creating an adaptive and synergistic interaction that continuously improves. We construct preference pairs by evaluating the impact of the weak model's contributions on collaborative performance. Then, we adopt direct preference optimization to align the weak model. This data-driven strategy allows us to amplify beneficial contributions from the weak model and minimize detrimental ones, thereby fostering a mutually beneficial interaction.
The pesudo codes of training and inference are Algorithm \ref{alg:training} and Algorithm \ref{alg:inference} in the appendix. 

\subsection{Problem Setup}

We propose a collaborative approach that leverages both weak and strong models to tackle diverse reasoning tasks. These tasks require domain-specific knowledge, problem-solving skills, and strong general capabilities such as reasoning, comprehension, and calculation. To address these tasks, we employ a \textbf{weak model} (e.g., Llama2-7b), denoted as $\pi_w$. This relatively small, cost-efficient model is a white-box system that can be fine-tuned for specific domains to acquire task-relevant knowledge. Alongside this, we utilize a \textbf{strong model} (e.g., GPT-4), referred to as $\pi_s$, a black-box model with fixed internal parameters. Although it has limited access to specific knowledge or proprietary data, the strong model excels in general reasoning.

Given a user query $x$ from a target task, our objective is to enhance the overall inference capability by utilizing the complementary strengths of $\pi_w$ and $\pi_s$. The inference process is formulated as:
\begin{equation}
y^* = \mathcal{F} \big( \pi_w \circ x, \, \pi_s \circ x, \, x \big) \quad \forall \, x \in X, \nonumber
\end{equation}
where $y^*$ represents the final output for the query $x$, and $\mathcal{F}$ is the mechanism that integrates the domain-specific expertise of $\pi_w$ with the general reasoning capability of $\pi_s$, for improved task performance.

\subsection{Supervised Fine-tuning of the Weak Model}
The weak model $\pi_w$ is initially fine-tuned on a task-specific training dataset, $\mathcal{D}_{\text{SFT}} = \{(x, \hat{y})\}$, where each query $x$ has a corresponding ground truth $\hat{y}$. The goal of this fine-tuning is to adapt $\pi_w$ to the specific task by learning from these examples. This is achieved by optimizing the following objective:
\begin{equation} \label{eqa:sft}
\pi_{\theta}^{\text{SFT}} = \arg \min_{\theta} \, \mathcal{L}_{\text{SFT}} \left( \pi_{\theta}; \, \mathcal{D}_{\text{SFT}} \right), 
\end{equation}
where $\pi_{\theta}^{\text{SFT}}$ is the policy after fine-tuning, and $\mathcal{L}_{\text{SFT}}$ is the supervised loss function to minimize the negative log-likelihood:
\begin{equation}\label{equ:sft}
\mathcal{L}_{\text{SFT}}(\pi_{\theta}) = - \mathbb{E}_{(x, y) \sim \mathcal{D}} \left[ \log \pi_{\theta}(y \mid x) \right] \nonumber
\end{equation}
This optimization allows the weak model to specialize in the task domain, preparing it for effective collaboration with the strong model.

\begin{figure}[!tbp]
    \centering    
    \includegraphics[width=0.48\textwidth]{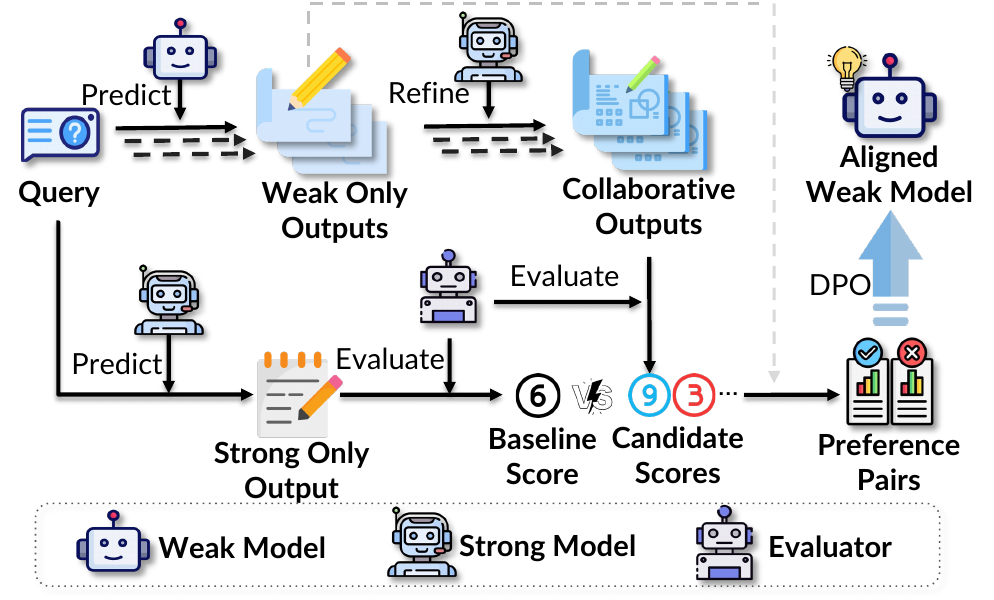}
    \caption{Aligning the Weak Model with Strong Model Feedback, including preference data construction and preference tuning. }
    \label{fig:model}
\end{figure}

\subsection{Aligning the Weak Model with Strong Model Feedback}

This subsection describes how to align the weak model with feedback from the strong model. Preference triplets are constructed by comparing the outputs produced solely by the strong model with those generated in collaboration with the weak model. An external evaluator scores these outputs based on reasoning coherence and alignment with the ground truth, identifying instances where the weak model’s contributions improve the final result. These triplets are then used to fine-tune the weak model through preference optimization, aligning it with the strong model's preferences to facilitate better collaboration.

\subsubsection{Preference Feedback from the Strong Model}

Given a set of training data, $\{(x, \hat{y})\}$, where $x$ is the query and $\hat{y}$ the groundtruth, our goal is to construct preference triplets $(x, y_+, y_-)$, where $y_+$ and $y_-$ represent the preferred and non-preferred outputs of the weak model. These triplets indicate whether the weak model's output enhances the final result in its collaboration with the strong model.

To construct these preference triplets, we introduce two generation scenarios:
\begin{itemize}[leftmargin=*]
    \item \textbf{Strong Model Only}: The query $x$ is directly fed into the strong model, which generates an explanation and a final output using a chain-of-thought (CoT) prompt. This approach helps the model break down complex tasks into intermediate reasoning steps. The resulting output is denoted as $z \sim \pi_s(z \mid x)$.
    \item \textbf{Weak-Strong Model Collaboration}: The query $x$ is first processed by the weak model to produce an explanation and an initial result, $y \sim \pi_w(y \mid x)$. This output, along with the original query, is then passed to the strong model for refinement, resulting in the final response $y^* \sim \pi_s(y^* \mid y)$. Here, the weak model's explanation may contain knowledge-intensive information that the strong model analyzes to detect potential flaws or gaps in reasoning.
\end{itemize}

\paragraph{Preference Evaluation}

To assess the output quality, we introduce an external evaluator, $E(y, x)$, which is a large language model with strong general capabilities (e.g., GPT-4). While various models can serve as the evaluator, using the same large language model as the strong model ensures consistency in reflecting the strong model's preferences. The evaluator scores the outputs based on a manually defined rubric: (1) Coherence of reasoning logic: whether the explanation is logically sound. (2) Consistency with ground truth: how closely the final result aligns with the ground truth. 

The evaluator $E$ assigns a fine-grained score to each output, providing a nuanced assessment of both the reasoning process and the final result. This model-based evaluation approach is preferred over traditional metrics like BLEU or ROUGE, as it captures not just surface similarity but also the depth of reasoning and logical coherence.

\paragraph{Preference Data Construction}

For each query $x$, we construct the preference triplet $(x, y_+, y_-)$ by comparing the evaluation scores of the strong model's output, $z \sim \pi_s(z \mid x)$, and the collaborative output, $\pi_s \circ y$. The preference is determined by the difference:
\[
\Delta = E(\pi_s \circ y, x) - E(z, x).
\]
If $\Delta > 0$, the weak model's contribution is deemed beneficial, and its output $y$ is selected as the positive response $y_+$. Conversely, if $\Delta \leq 0$, $y$ is designated as the negative response $y_-$.
The preference data is formalized using two conditional probability distributions over the weak model's outputs:
\begin{equation}
    \begin{aligned}
        &p_+(y_+ \mid z, x) = \\
        &\frac{\pi_{w}(y_+ \mid x) \, \mathds{1} \left\{ E(\pi_s \circ y_+, x) > E(z, x) \right\}}
        {\int \pi_{w}(y \mid x) \, \mathds{1} \left\{ E(\pi_s \circ y, x) > E(z, x) \right\} \, dy}. \nonumber
    \end{aligned}
\end{equation}

\begin{equation}
    \begin{aligned}
        &p_-(y_- \mid z, x) =  \\
        &\frac{\pi_w(y_- \mid x) \, \mathds{1} \left\{ E(\pi_s \circ y_-, x) \leq E(z, x) \right\}}{\int \pi_w(y \mid x) \, \mathds{1} \left\{ E(\pi_s \circ y, x) \leq E(z, x) \right\} \, dy}. \nonumber
    \end{aligned}
\end{equation}

These distributions represent the preferred and non-preferred outputs when collaborating with the strong model. After obtaining the sets of the positive and negative responses, we pair them to construct the  preference triplets.

\subsubsection{Preference Tuning for the Weak Model}\label{sec:objective}

Using the constructed preference triplets $\mathcal{D}_{\text{PT}} = \{(x, y_+, y_-)\}$, we fine-tune the weak model $\pi_w$ to align its outputs with those that are preferred in collaboration with the strong model. We employ Direct Preference Optimization (DPO)~\citep{DBLP:conf/nips/RafailovSMMEF23} to adjust the weak model's policy $\pi_w$. The DPO objective is formulated as :
\begin{equation}\label{equ:obj}
    \begin{split}
        &\mathcal{L}_{\text{DPO}} = \min_{\pi_w^*} 
        -\mathbb{E}_{\substack{x, \, z \sim \pi_s(z \mid x), \\ 
        y_+ \sim p_w(\cdot \mid z, x), \\ 
        y_- \sim p_-(\cdot \mid z, x)}} 
        \Bigg[ \\ 
        &\log \sigma \Bigg( \alpha \log \frac{\pi_w^*(y_+ \mid x)}{\pi_w(y_+ \mid x)} 
        - \alpha \log \frac{\pi_w^*(y_- \mid x)}{\pi_w(y_- \mid x)} 
        \Bigg) \Bigg] \nonumber \\
    \end{split}
\end{equation}
where $\sigma(\cdot)$ is the logistic sigmoid function, and $\alpha$ is a scaling parameter. By optimizing this objective, we encourage the weak model to generate outputs that lead to higher scores when refined by the strong model.

The overall objective is to find the optimal policy:
\begin{equation}\label{eqa:pt}
\pi_w^* = \arg\min \mathcal{L}_{\text{DPO}}(\pi_w; \pi_w^{\text{SFT}}; \mathcal{D}_{\text{PT}}), 
\end{equation}
where $\pi_w^*$ is the optimal policy aligned with the strong model's preferences, and $\pi_w^{\text{SFT}}$ is the reference weak model obtained through supervised fine-tuning.

\subsection{Collaborative Inference}
During inference, the input query $x$ is first processed by the weak model $\pi_w^*$ to generate an initial output. This output, along with the original query, is then passed to the strong model $\pi_s$ for refinement, resulting in the final answer:
$$
y^* = \pi_s \circ (x, \pi_w^* \circ x).
$$
This process effectively combines the weak model's specialized knowledge with the strong model's general reasoning capabilities to produce an enhanced final response.

\subsection{Theoretical Insight} 

In this section, we build on the methodology discussed earlier to present a formal theoretical analysis of how the proposed preference-based alignment affects the weak model's behavior and performance. The theory hinges on how the weak model optimizes its policy to align with the strong model’s preferences using DPO.

For simplicity, we assume that the evaluator scores for the strong model's outputs are constant for all $z$, i.e. $E(z, x) = p(x)$ for all $z$ when given $x$. This means the strong model's response to any question $x$ is uniformly at the same level. Under this assumption, we aim to understand the behavior of the newly optimized weak model $\pi_{w}^*$.

Regarding the optimization objective $\mathcal{L}_{\text{DPO}}$ in Section \ref{sec:objective}, the key aspect is that the positive (\( p_+(\cdot | z, x) \)) and negative (\( p_-(\cdot | z, x) \)) responses have disjoint support. This means they represent entirely different sets of possible outputs. As a result, the optimized weak model \( \pi_w^* \) allocates zero probability to any output \( y \) that results in an evaluator score \( E(\pi_s \circ y, x) \leq p(x) \). This finding implies:
\[
\pi_{w}^*(y \mid x) = 0,  \forall y \text{ such that } E(\pi_s \circ y, x) \leq p(x).
\]

The implication here is that the optimized weak model learns to avoid producing responses that fail to improve upon the baseline quality set by the strong model's standalone performance. Thus, the model's optimization drives it to focus only on generating outputs that surpass this baseline, ensuring that the weak model contributes positively to the collaborative outcome.

\renewcommand{\arraystretch}{1.3}
\begin{table*} 
    \centering
    \small
    \tabcolsep0.14 in
    \begin{NiceTabular}{llcccccc}
        \CodeBefore
        \Body
        \toprule
        \multirow{2}{*}{\textbf{Methods}} & \multirow{2}{*}{\textbf{Models}} & \multicolumn{2}{c}{\textbf{Counterfactuals}} & \multicolumn{2}{c}{\textbf{Medicine}} & \multicolumn{2}{c}{\textbf{Ethics}} \\
         & & \textbf{EM} & \textbf{F1} & \textbf{Acc.} & \textbf{F1} & \textbf{Acc.} & \textbf{F1} \\
        \midrule
        \multirow{3}{*}{\textbf{Weak Only}} & Llama-3-8B & 68.57 & 71.85 & 59.48 & 46.99 & 38.10 & 36.40 \\
        & Llama-3-8B (SFT) & 69.71 & 72.69 & 73.08 & 58.26 & 64.29 & 62.40 \\
        & Llama-3-8B (Self-Refine) & 70.52 & 72.99 & 60.71 & 47.07 & 35.89 & 35.06 \\
        \midrule
        \multirow{3}{*}{\textbf{Strong Only}} 
         & GPT-4 & 49.44 & 60.93 & 65.87 & 54.86 & 36.75 & 35.25 \\
         & GPT-4 (CoT) & 57.42 & 65.60 & 71.80 & 57.69 & 39.00 & 39.58 \\
         & GPT-4 (Self-Refine) & 61.54 & 68.83 & 72.95 & 59.10 & 38.16 & 37.87 \\
        \midrule
        \multirow{2}{*}{\textbf{RAG}} & SKR \citep{wang2023self} & 59.75 & 68.33 & 71.90 & 56.37 & 56.46 & 55.40 \\ 
         & FLARE \citep{DBLP:conf/emnlp/JiangXGSLDYCN23} & 62.07 & 70.59 & 72.40 & 58.89 & 55.27 & 54.97 \\
        \midrule
        \multirow{4}{*}{\textbf{Collaboration}} 
        & RLWF \citep{tao2024your} & 70.52 & 75.04 & 72.01 & 57.65 & 64.85 & 62.10 \\
        & RLAIF \citep{DBLP:conf/icml/0001PMMFLBHCRP24} & 71.69 & 72.17 & 71.91 & 57.55 & 62.74 & 59.65 \\
        & SuperICL \citep{DBLP:conf/acl/XuXWLZM24} & 68.85 & 74.82 & 73.64 & 58.33 & 66.18 & 63.86 \\
        & \textbf{CoWest (Ours)} & \textbf{75.85} & \textbf{77.34} & \textbf{75.10} & \textbf{60.13} & \textbf{68.33} & \textbf{65.61} \\
        \bottomrule
    \end{NiceTabular}
    \caption{Experiment results across three datasets. Results are reported as Exact Match (EM) and F1 scores for the Counterfacual dataset, Accuracy (Acc) and F1 for the Medical and Ethics datasets. Here, CoT is Chain of thought \citep{wei2022chain} while Self-Refine means iterative refining LLM with self-feedback from \citet{madaan2024self}.}
    \label{tab:exp}
\end{table*}

Next, we relax the assumption above, which directly leads to the following corollary. 

\textbf{Corollary 1:} Assuming the strong model's responses are not just uniform but also bounded below by some quality threshold: \( p(z) \leq E(z, x) \) for all \( z \), the newly optimized weak model \( \pi_w^*(x) \) will strictly avoid producing any response \( y \) for which the collaborative evaluation score fails to exceed the baseline:
\[
E(\pi_s \circ y, x) \leq p(x).
\]
The proof idea is exactly as the analysis above. In addition, this means that the weak model, through preference optimization, learns to consistently produce only those responses that align with or surpass the evaluator's expectations. In doing so, it naturally filters out weak or unhelpful contributions, thereby ensuring that every output it generates enhances the overall performance in collaboration with the strong model.

\section{Experiment}

\subsection{Experiment Setting}

\paragraph{Dataset} We evaluate our framework on three specialized datasets from different domains:
(1) \textbf{Counterfactuals}: IfQA \citep{yu-etal-2023-ifqa} is a counterfactual QA benchmark with questions based on hypothetical “if” clauses, requiring reasoning about imagined situations that may contradict factual knowledge.
(2) \textbf{Medicine}: MedMCQA \citep{pmlr-v174-pal22a} is a multiple-choice QA dataset of real-world medical exam questions, demanding in-depth language understanding and reasoning.
(3) \textbf{Ethics}: Prosocial-Dialog \citep{kim2022prosocialdialog} is a multi-turn English dialogue dataset covering diverse unethical content, with responses classified by safety levels.
More details can be found in Appendix \ref{sec:app_dataset}.

\paragraph{Evaluation Metrics} 
For IfQA, an open-ended question answering task, we use two commonly used metrics to evaluate the performance: exact match (EM) and F1 score following the setting of previous work \citep{sachan2023questions, yu-etal-2023-ifqa}.
For MedMCQA, a multi-choice question answering task, we use accuracy as the primary evaluation metric.  Additionally, we consider using macro-averaged F1 score to capture the model's performance across all answer categories.
For Prosocial-Dialog, a classification task, we utilize macro-F1 scores and accuracy as evaluation metrics to assess the model's capability in classifying responses based on prosocial behaviors. 

\begin{figure*}[!tbp]
    \centering
    \includegraphics[width=0.9\linewidth]{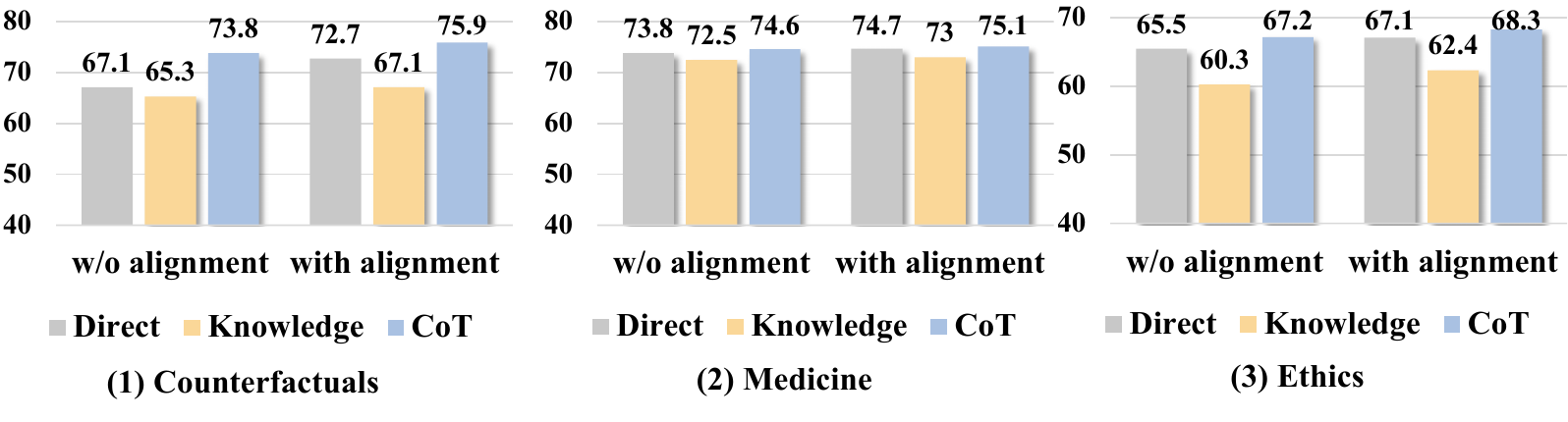}
    \vspace{-10pt}
    \caption{Analysis of different interaction strategies between weak-strong models in \ours indicates alignment boosts the performance obviously. We report EM  for Counterfactuals and Accuracy for Medicine and Ethics.}
    \label{fig:analysis}
\end{figure*}

\paragraph{Implementation Details} 
In our experiments, we utilize two models: the weak model, LLaMA3-8B \citep{dubey2024llama}, and the strong model, GPT-4-0613 \citep{achiam2023gpt} for Counterfactuals and Medicine and GPT-3.5-Turbo for Ethics. For the evaluator, we use the same model as the strong model. For the fine-tuning of the weak model, we employ Low-Rank Adaptation (LoRA) \citep{DBLP:journals/corr/abs-2106-09685}. We generate 2,000 pieces of data for IFQA and 5,000 pieces for MedMCQA and Prosocial-Dialog.
More details of model training and prompt design can be found in Appendix \ref{sec:app_imple}.

\paragraph{Baselines} The baselines include the following categories: 
(1) \textbf{Weak Model}: We employ both weak and strong models alone. For weak models, we include Llama-3-8B \citep{dubey2024llama}, Llama-3-8B-SFT and Llama-3-8B with Self-Refine \citep{madaan2024self}. 
(2) \textbf{Strong Model}: we test zero-shot GPT-3.5-Turbo-0613 and GPT-4-0613, including their variants with chain-of-thought \citep{wei2022chain} and Self-Refine \citep{madaan2024self}.
(3) \textbf{Retrieval-Augmented Generation}: We involve two methods, SKR \citep{wang2023self} and FLARE  \citep{DBLP:conf/emnlp/JiangXGSLDYCN23}. For fair comparison, we adopt GPT-4 as their backbone.
(4) \textbf{Model Collaboration}: There are three prior methods, including SuperICL \citep{DBLP:conf/acl/XuXWLZM24}, RLWF \citep{tao2024your} and RLAIF \citep{DBLP:conf/icml/0001PMMFLBHCRP24}.
Please find more details in Appendix \ref{sec:app_baseline}.

\begin{figure}[!tbp]
    \centering
    \includegraphics[width=\linewidth]{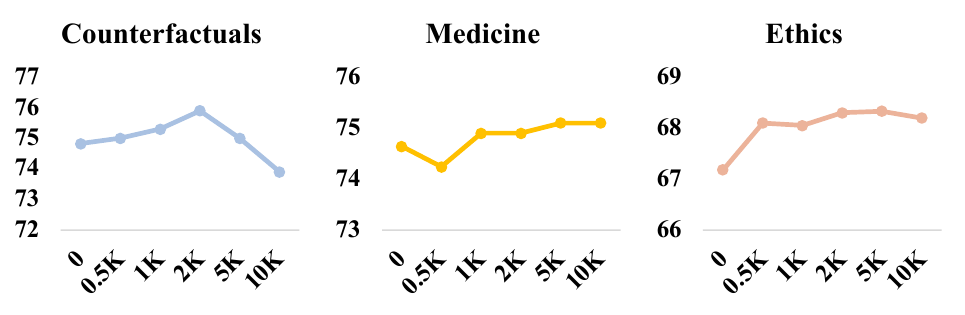}
    \caption{Analysis of different sizes of training data for preference tuning.}
    \label{fig:analysis3}
\end{figure}

\subsection{Experiment Result}
According to the evaluation results in Table \ref{tab:exp}, our major observation is \textbf{weak-strong model collaboration leads to substantial improvements over single models}. Our collaborative framework, \ours, demonstrates clear performance gains across all datasets when compared to the single models. For instance, \ours  \ improves over the best-performing single model (LLaMA3-8B after finetuning) by a significant margin, particularly on the IfQA and Prosocial-Dialog datasets. This underscores the effectiveness of combining a specialized weak model with a general-purpose strong model, allowing each to compensate for the other's limitations. 
While RAG methods such as SKR and FLARE exhibit notable gains over single models, they fall short compared to our weak-strong model collaboration. Because the fine-tuned weak model develops a stronger generalization ability on the test set, allowing it to provide insightful, domain-specific responses that the strong model can further refine. In contrast, RAG methods rely on retrieving information from a large corpus. It lacks the adaptability needed for specialized tasks.

\subsection{Analysis}

\begin{figure}[!tbp]
    \centering
    \includegraphics[width=\linewidth]{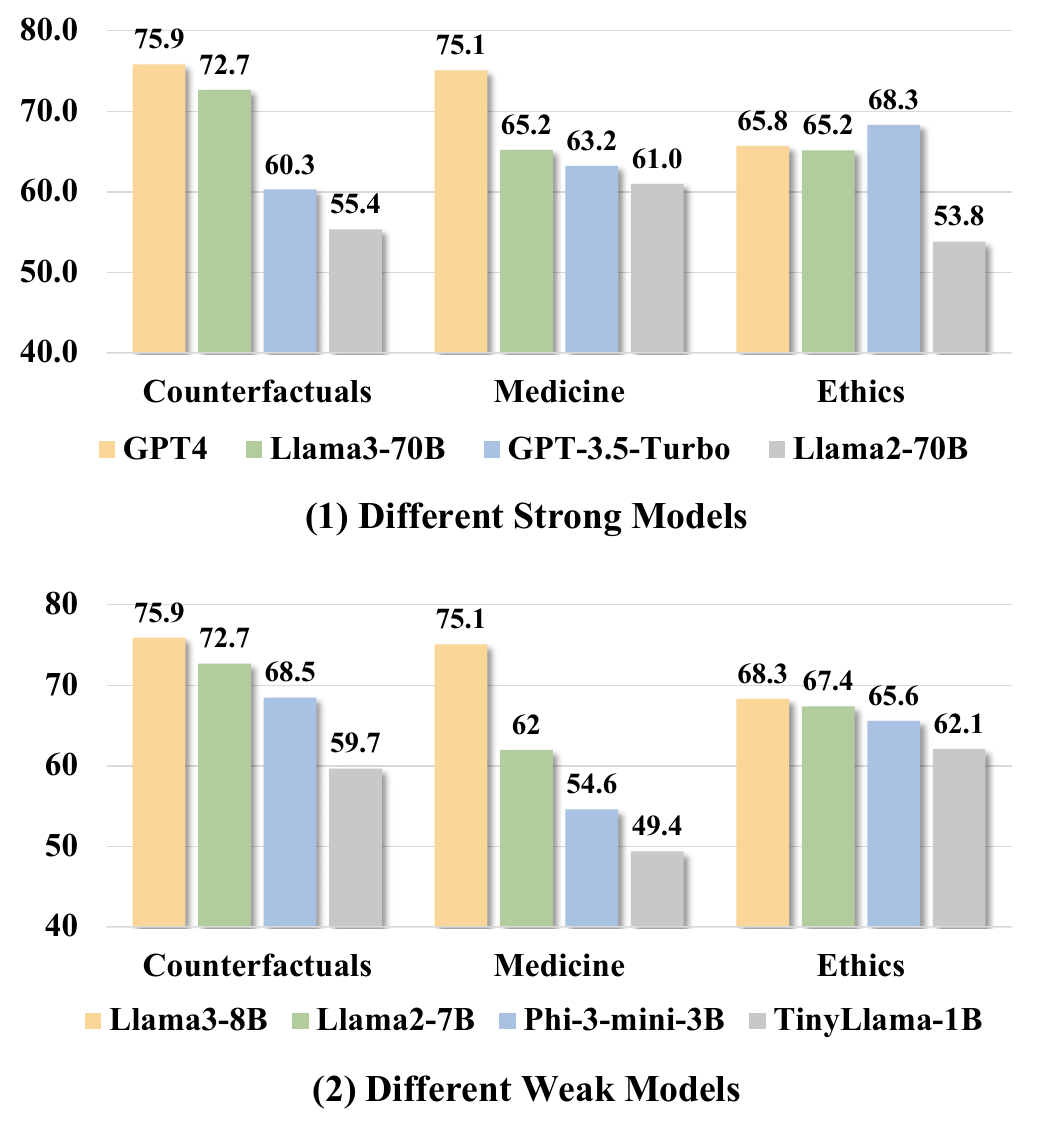}
    \caption{Analysis of adopting different weak or strong models in \ours \ while fixing the other.}
    \label{fig:analysis2}
\end{figure}

\paragraph{Interaction strategies between weak-strong models}

In our experiments, we examine two key interaction strategies between weak and strong models: (1) W/o Alignment, where the weak model generates initial responses that the strong model then refines without any model alignment, and (2) With Alignment, which involves fine-tuning the weak model based on the strong model's preferences. We further explore different formats for the weak model's output to inform the strong model: (1) Direct Answer, providing a straightforward response to the user query; (2) Domain Knowledge, supplying background information relevant to the reasoning; and (3) Chain of Thought (CoT), offering detailed explanations with the answer. By combining these two interaction strategies with the three formats, we assess each combination's effectiveness in handling specialized tasks. We report the EM scores for Counterfactuals and the accuracy scores for Medicine and Ethics.

As shown in Figure \ref{fig:analysis2}, our experiments clearly demonstrate the effectiveness of model alignment across all three datasets comparing the results with and without alignment.
Particularly, the Chain of Thought (CoT) format stands out as the most advantageous, surpassing both Direct Answer and Domain Knowledge formats. Its detailed reasoning path significantly assists the strong model in handling complex queries, as seen in its stronger performance on the ethics and counterfactual datasets—both of which demand advanced reasoning. In contrast, the medicine dataset, requiring substantial domain knowledge, shows less variation across interaction strategies, suggesting that task-specific expertise can outweigh interaction style when the knowledge requirement is paramount.

\paragraph{Impact of different strong models: General capabilities enhance problem-solving.}
In this setup, we standardized the strong model for specific domains. Llama-3-8B served as the weak model across all datasets, allowing us to evaluate the performance of different strong models—GPT-4, Llama-3-70B \citep{dubey2024llama}, GPT-3.5-Turbo, and Llama-2-70B \citep{touvron2023llama}—across various domains.
According to the experiment results in Figure \ref{fig:analysis2}, the strong model GPT-4, when engaged in the domain of Counterfactuals, exhibits the highest accuracy at 75.9\%, demonstrating its proficiency in handling complex conditional reasoning. Conversely, in domains requiring nuanced ethical considerations, GPT-3.5-Turbo outperforms other models with an accuracy of 68.3\%. This indicates that the effectiveness of strong models is highly domain-dependent, where their inherent strengths can enhance overall performance.

\paragraph{Impact of different weak models: Foundation and adaptability are key.}

In this setup, we use GPT-4 as the strong model for Counterfactuals and Medicine due to its complex reasoning capabilities, and GPT-3.5-Turbo was used for Ethics to handle nuanced moral dilemmas. The involved weak models include Llama-3-8B \citep{dubey2024llama}, Llama-2-7B \citep{touvron2023llama}, Phi-3-mini-3B \citep{abdin2024phi}, and TinyLlama-1B \citep{zhang2024tinyllama}.
According to the experiment results in Figure \ref{fig:analysis}, the selection and performance of weak models, such as Llama-3-8B and Llama-2-7B, clearly show a superior handling of tasks across all domains compared to smaller models like Phi-3-mini-3B and TinyLlama-1B. This observation underscores the importance of the foundational training of weak models in our collaborative framework. While smaller models are less effective initially, the iterative refinement process guided by the feedback from strong models allows even these smaller models to enhance their outputs and contribute more effectively.

\paragraph{Impact of different sizes of training data for preference tuning}
As shown in Figure \ref{fig:analysis}, increasing training data generally improves performance, but its impact varies across datasets. For Counterfactuals, where the original dataset contains only 2K samples, expanding preference tuning data requires repeated sampling, potentially lowering quality. This likely explains why performance peaks at 1K, with larger datasets introducing redundancy. In contrast, Medicine and Ethics have larger original datasets, allowing for selective expansion while maintaining quality, leading to continued gains up to 2K samples. \textbf{These results highlight the trade-off between data quantity and quality}—larger datasets help, but only when high-quality samples can be ensured.

\section{Conclusion}
In conclusion, our research has demonstrated the significant potential of leveraging a collaborative framework between weak and strong models to address specialized tasks effectively. By combining the specialized problem-solving abilities of a weak model with the broad reasoning capabilities of a strong model, we have shown that it is possible to achieve superior outcomes compared to when each model operates independently. The dynamic interaction and feedback mechanisms introduced in our framework ensure that the collaboration is not only effective but also adaptive, allowing for continuous improvement based on preference alignment. 

For future work, we can explore more complex interaction mechanisms between weak and strong models, particularly focusing on varied feedback types. Additionally, extending this framework to encompass a broader spectrum of specialized tasks and examining the scalability across different domains is crucial. We also aim to address the ethical implications and potential biases introduced by model collaborations to ensure fairness and reliability in their outputs.

\section*{Limitations}

\paragraph{Single-Iteration Feedback} In this study, we primarily concentrated on incorporating a single feedback iteration between the weak and strong models. While this choice simplified our experimental setup and allowed for initial insights, it does not capture the full potential of iterative refinement. Future work should explore multiple rounds of feedback to determine how repeated interactions could further enhance model performance and adaptation.

\paragraph{Restricted Model Families} Our experiments focused on GPT-related and Llama-related model. Although these models are representative of strong and weak reasoning capabilities, respectively, our findings may not generalize to other model families or architectures. Examining whether similar collaborative benefits can be observed with additional models remains an open avenue for future research.

\paragraph{Computational Overhead Analysis} While we highlight the promise of feedback-driven collaboration, a detailed examination of the computational cost associated with these iterative interactions was beyond the scope of this work. Analyzing overhead—such as the number of passes required, memory utilization, and latency—will be essential for understanding the practical limits of deploying this framework at scale.


\bibliography{0_main}

\begin{thebibliography}{50}
\providecommand{\natexlab}[1]{#1}

\bibitem[{Abdin et~al.(2024)Abdin, Jacobs, Awan, Aneja, Awadallah, Awadalla,
  Bach, Bahree, Bakhtiari, Behl et~al.}]{abdin2024phi}
Marah Abdin, Sam~Ade Jacobs, Ammar~Ahmad Awan, Jyoti Aneja, Ahmed Awadallah,
  Hany Awadalla, Nguyen Bach, Amit Bahree, Arash Bakhtiari, Harkirat Behl,
  et~al. 2024.
\newblock Phi-3 technical report: A highly capable language model locally on
  your phone.
\newblock \emph{arXiv preprint arXiv:2404.14219}.

\bibitem[{Achiam et~al.(2023)Achiam, Adler, Agarwal, Ahmad, Akkaya, Aleman,
  Almeida, Altenschmidt, Altman, Anadkat et~al.}]{achiam2023gpt}
Josh Achiam, Steven Adler, Sandhini Agarwal, Lama Ahmad, Ilge Akkaya,
  Florencia~Leoni Aleman, Diogo Almeida, Janko Altenschmidt, Sam Altman,
  Shyamal Anadkat, et~al. 2023.
\newblock Gpt-4 technical report.
\newblock \emph{arXiv preprint arXiv:2303.08774}.

\bibitem[{Burns et~al.(2024)Burns, Izmailov, Kirchner, Baker, Gao,
  Aschenbrenner, Chen, Ecoffet, Joglekar, Leike, Sutskever, and
  Wu}]{DBLP:conf/icml/BurnsIKBGACEJLS24}
Collin Burns, Pavel Izmailov, Jan~Hendrik Kirchner, Bowen Baker, Leo Gao,
  Leopold Aschenbrenner, Yining Chen, Adrien Ecoffet, Manas Joglekar, Jan
  Leike, Ilya Sutskever, and Jeffrey Wu. 2024.
\newblock \href {https://openreview.net/forum?id=ghNRg2mEgN} {Weak-to-strong
  generalization: Eliciting strong capabilities with weak supervision}.
\newblock In \emph{Forty-first International Conference on Machine Learning,
  {ICML} 2024, Vienna, Austria, July 21-27, 2024}. OpenReview.net.

\bibitem[{Chang et~al.(2024)Chang, Wang, Wang, Wu, Yang, Zhu, Chen, Yi, Wang,
  Wang et~al.}]{chang2024survey}
Yupeng Chang, Xu~Wang, Jindong Wang, Yuan Wu, Linyi Yang, Kaijie Zhu, Hao Chen,
  Xiaoyuan Yi, Cunxiang Wang, Yidong Wang, et~al. 2024.
\newblock A survey on evaluation of large language models.
\newblock \emph{ACM Transactions on Intelligent Systems and Technology},
  15(3):1--45.

\bibitem[{Charikar et~al.(2024)Charikar, Pabbaraju, and
  Shiragur}]{DBLP:journals/corr/abs-2405-15116}
Moses Charikar, Chirag Pabbaraju, and Kirankumar Shiragur. 2024.
\newblock \href {https://doi.org/10.48550/ARXIV.2405.15116} {Quantifying the
  gain in weak-to-strong generalization}.
\newblock \emph{CoRR}, abs/2405.15116.

\bibitem[{Deng and Raffel(2023)}]{DBLP:conf/emnlp/DengR23}
Haikang Deng and Colin Raffel. 2023.
\newblock \href {https://doi.org/10.18653/V1/2023.EMNLP-MAIN.721}
  {Reward-augmented decoding: Efficient controlled text generation with a
  unidirectional reward model}.
\newblock In \emph{Proceedings of the 2023 Conference on Empirical Methods in
  Natural Language Processing, {EMNLP} 2023, Singapore, December 6-10, 2023},
  pages 11781--11791. Association for Computational Linguistics.

\bibitem[{Dubey et~al.(2024)Dubey, Jauhri, Pandey, Kadian, Al-Dahle, Letman,
  Mathur, Schelten, Yang, Fan et~al.}]{dubey2024llama}
Abhimanyu Dubey, Abhinav Jauhri, Abhinav Pandey, Abhishek Kadian, Ahmad
  Al-Dahle, Aiesha Letman, Akhil Mathur, Alan Schelten, Amy Yang, Angela Fan,
  et~al. 2024.
\newblock The llama 3 herd of models.
\newblock \emph{arXiv preprint arXiv:2407.21783}.

\bibitem[{Fedus et~al.(2022)Fedus, Zoph, and
  Shazeer}]{DBLP:journals/jmlr/FedusZS22}
William Fedus, Barret Zoph, and Noam Shazeer. 2022.
\newblock \href {https://jmlr.org/papers/v23/21-0998.html} {Switch
  transformers: Scaling to trillion parameter models with simple and efficient
  sparsity}.
\newblock \emph{J. Mach. Learn. Res.}, 23:120:1--120:39.

\bibitem[{Fu et~al.(2023)Fu, Peng, Ou, Sabharwal, and
  Khot}]{fu2023specializing}
Yao Fu, Hao Peng, Litu Ou, Ashish Sabharwal, and Tushar Khot. 2023.
\newblock Specializing smaller language models towards multi-step reasoning.
\newblock In \emph{International Conference on Machine Learning}, pages
  10421--10430. PMLR.

\bibitem[{Guo and Yang(2024)}]{guo2024improving}
Yue Guo and Yi~Yang. 2024.
\newblock Improving weak-to-strong generalization with reliability-aware
  alignment.
\newblock \emph{arXiv preprint arXiv:2406.19032}.

\bibitem[{Guu et~al.(2020)Guu, Lee, Tung, Pasupat, and
  Chang}]{DBLP:conf/icml/GuuLTPC20}
Kelvin Guu, Kenton Lee, Zora Tung, Panupong Pasupat, and Ming{-}Wei Chang.
  2020.
\newblock \href {http://proceedings.mlr.press/v119/guu20a.html} {Retrieval
  augmented language model pre-training}.
\newblock In \emph{Proceedings of the 37th International Conference on Machine
  Learning, {ICML} 2020, 13-18 July 2020, Virtual Event}, volume 119 of
  \emph{Proceedings of Machine Learning Research}, pages 3929--3938. {PMLR}.

\bibitem[{Hu et~al.(2021)Hu, Shen, Wallis, Allen{-}Zhu, Li, Wang, and
  Chen}]{DBLP:journals/corr/abs-2106-09685}
Edward~J. Hu, Yelong Shen, Phillip Wallis, Zeyuan Allen{-}Zhu, Yuanzhi Li,
  Shean Wang, and Weizhu Chen. 2021.
\newblock \href {https://arxiv.org/abs/2106.09685} {Lora: Low-rank adaptation
  of large language models}.
\newblock \emph{CoRR}, abs/2106.09685.

\bibitem[{Izacard et~al.(2022)Izacard, Lewis, Lomeli, Hosseini, Petroni,
  Schick, Dwivedi{-}Yu, Joulin, Riedel, and
  Grave}]{DBLP:journals/corr/abs-2208-03299}
Gautier Izacard, Patrick S.~H. Lewis, Maria Lomeli, Lucas Hosseini, Fabio
  Petroni, Timo Schick, Jane Dwivedi{-}Yu, Armand Joulin, Sebastian Riedel, and
  Edouard Grave. 2022.
\newblock \href {https://doi.org/10.48550/ARXIV.2208.03299} {Few-shot learning
  with retrieval augmented language models}.
\newblock \emph{CoRR}, abs/2208.03299.

\bibitem[{Ji et~al.(2024)Ji, Chen, Lou, Hong, Zhang, Pan, Dai, and
  Yang}]{ji2024aligner}
Jiaming Ji, Boyuan Chen, Hantao Lou, Donghai Hong, Borong Zhang, Xuehai Pan,
  Juntao Dai, and Yaodong Yang. 2024.
\newblock Aligner: Achieving efficient alignment through weak-to-strong
  correction.
\newblock \emph{arXiv preprint arXiv:2402.02416}.

\bibitem[{Jiang et~al.(2024)Jiang, Sablayrolles, Roux, Mensch, Savary, Bamford,
  Chaplot, de~Las~Casas, Hanna, Bressand, Lengyel, Bour, Lample, Lavaud,
  Saulnier, Lachaux, Stock, Subramanian, Yang, Antoniak, Scao, Gervet, Lavril,
  Wang, Lacroix, and Sayed}]{DBLP:journals/corr/abs-2401-04088}
Albert~Q. Jiang, Alexandre Sablayrolles, Antoine Roux, Arthur Mensch, Blanche
  Savary, Chris Bamford, Devendra~Singh Chaplot, Diego de~Las~Casas, Emma~Bou
  Hanna, Florian Bressand, Gianna Lengyel, Guillaume Bour, Guillaume Lample,
  L{\'{e}}lio~Renard Lavaud, Lucile Saulnier, Marie{-}Anne Lachaux, Pierre
  Stock, Sandeep Subramanian, Sophia Yang, Szymon Antoniak, Teven~Le Scao,
  Th{\'{e}}ophile Gervet, Thibaut Lavril, Thomas Wang, Timoth{\'{e}}e Lacroix,
  and William~El Sayed. 2024.
\newblock \href {https://doi.org/10.48550/ARXIV.2401.04088} {Mixtral of
  experts}.
\newblock \emph{CoRR}, abs/2401.04088.

\bibitem[{Jiang et~al.(2023{\natexlab{a}})Jiang, Ren, and
  Lin}]{DBLP:conf/acl/Jiang0L23}
Dongfu Jiang, Xiang Ren, and Bill~Yuchen Lin. 2023{\natexlab{a}}.
\newblock \href {https://doi.org/10.18653/V1/2023.ACL-LONG.792} {Llm-blender:
  Ensembling large language models with pairwise ranking and generative
  fusion}.
\newblock In \emph{Proceedings of the 61st Annual Meeting of the Association
  for Computational Linguistics (Volume 1: Long Papers), {ACL} 2023, Toronto,
  Canada, July 9-14, 2023}, pages 14165--14178. Association for Computational
  Linguistics.

\bibitem[{Jiang et~al.(2023{\natexlab{b}})Jiang, Xu, Gao, Sun, Liu,
  Dwivedi{-}Yu, Yang, Callan, and Neubig}]{DBLP:conf/emnlp/JiangXGSLDYCN23}
Zhengbao Jiang, Frank~F. Xu, Luyu Gao, Zhiqing Sun, Qian Liu, Jane
  Dwivedi{-}Yu, Yiming Yang, Jamie Callan, and Graham Neubig.
  2023{\natexlab{b}}.
\newblock \href {https://doi.org/10.18653/V1/2023.EMNLP-MAIN.495} {Active
  retrieval augmented generation}.
\newblock In \emph{Proceedings of the 2023 Conference on Empirical Methods in
  Natural Language Processing, {EMNLP} 2023, Singapore, December 6-10, 2023},
  pages 7969--7992. Association for Computational Linguistics.

\bibitem[{Juneja et~al.(2023)Juneja, Dutta, Chakrabarti, Manchanda, and
  Chakraborty}]{juneja-etal-2023-small}
Gurusha Juneja, Subhabrata Dutta, Soumen Chakrabarti, Sunny Manchanda, and
  Tanmoy Chakraborty. 2023.
\newblock \href {https://doi.org/10.18653/v1/2023.emnlp-main.225} {Small
  language models fine-tuned to coordinate larger language models improve
  complex reasoning}.
\newblock In \emph{Proceedings of the 2023 Conference on Empirical Methods in
  Natural Language Processing}, pages 3675--3691, Singapore. Association for
  Computational Linguistics.

\bibitem[{Kim et~al.(2022)Kim, Yu, Jiang, Lu, Khashabi, Kim, Choi, and
  Sap}]{kim2022prosocialdialog}
Hyunwoo Kim, Youngjae Yu, Liwei Jiang, Ximing Lu, Daniel Khashabi, Gunhee Kim,
  Yejin Choi, and Maarten Sap. 2022.
\newblock Prosocialdialog: A prosocial backbone for conversational agents.
\newblock In \emph{EMNLP}.

\bibitem[{Kojima et~al.(2022)Kojima, Gu, Reid, Matsuo, and
  Iwasawa}]{kojima2022large}
Takeshi Kojima, Shixiang~Shane Gu, Machel Reid, Yutaka Matsuo, and Yusuke
  Iwasawa. 2022.
\newblock Large language models are zero-shot reasoners.
\newblock \emph{Advances in neural information processing systems},
  35:22199--22213.

\bibitem[{Lee et~al.(2024)Lee, Phatale, Mansoor, Mesnard, Ferret, Lu, Bishop,
  Hall, Carbune, Rastogi, and Prakash}]{DBLP:conf/icml/0001PMMFLBHCRP24}
Harrison Lee, Samrat Phatale, Hassan Mansoor, Thomas Mesnard, Johan Ferret,
  Kellie Lu, Colton Bishop, Ethan Hall, Victor Carbune, Abhinav Rastogi, and
  Sushant Prakash. 2024.
\newblock \href {https://openreview.net/forum?id=uydQ2W41KO} {{RLAIF} vs.
  {RLHF:} scaling reinforcement learning from human feedback with {AI}
  feedback}.
\newblock In \emph{Forty-first International Conference on Machine Learning,
  {ICML} 2024, Vienna, Austria, July 21-27, 2024}. OpenReview.net.

\bibitem[{Lewkowycz et~al.(2022)Lewkowycz, Andreassen, Dohan, Dyer,
  Michalewski, Ramasesh, Slone, Anil, Schlag, Gutman-Solo
  et~al.}]{lewkowycz2022solving}
Aitor Lewkowycz, Anders Andreassen, David Dohan, Ethan Dyer, Henryk
  Michalewski, Vinay Ramasesh, Ambrose Slone, Cem Anil, Imanol Schlag, Theo
  Gutman-Solo, et~al. 2022.
\newblock Solving quantitative reasoning problems with language models.
\newblock \emph{Advances in Neural Information Processing Systems},
  35:3843--3857.

\bibitem[{Liu et~al.(2024)Liu, Yang, Zhang, Hu, Li, Yan, Zhang, Huang, and
  Liu}]{liu2024panda}
An~Liu, Zonghan Yang, Zhenhe Zhang, Qingyuan Hu, Peng Li, Ming Yan, Ji~Zhang,
  Fei Huang, and Yang Liu. 2024.
\newblock Panda: Preference adaptation for enhancing domain-specific abilities
  of llms.
\newblock \emph{arXiv preprint arXiv:2402.12835}.

\bibitem[{Lu et~al.(2024)Lu, Pang, Xiao, Zhu, Xia, and Zhang}]{lu2024merge}
Jinliang Lu, Ziliang Pang, Min Xiao, Yaochen Zhu, Rui Xia, and Jiajun Zhang.
  2024.
\newblock Merge, ensemble, and cooperate! a survey on collaborative strategies
  in the era of large language models.
\newblock \emph{arXiv preprint arXiv:2407.06089}.

\bibitem[{Madaan et~al.(2024)Madaan, Tandon, Gupta, Hallinan, Gao, Wiegreffe,
  Alon, Dziri, Prabhumoye, Yang et~al.}]{madaan2024self}
Aman Madaan, Niket Tandon, Prakhar Gupta, Skyler Hallinan, Luyu Gao, Sarah
  Wiegreffe, Uri Alon, Nouha Dziri, Shrimai Prabhumoye, Yiming Yang, et~al.
  2024.
\newblock Self-refine: Iterative refinement with self-feedback.
\newblock \emph{Advances in Neural Information Processing Systems}, 36.

\bibitem[{O'Brien and Lewis(2023)}]{DBLP:journals/corr/abs-2309-09117}
Sean O'Brien and Mike Lewis. 2023.
\newblock \href {https://doi.org/10.48550/ARXIV.2309.09117} {Contrastive
  decoding improves reasoning in large language models}.
\newblock \emph{CoRR}, abs/2309.09117.

\bibitem[{Pal et~al.(2022)Pal, Umapathi, and Sankarasubbu}]{pmlr-v174-pal22a}
Ankit Pal, Logesh~Kumar Umapathi, and Malaikannan Sankarasubbu. 2022.
\newblock \href {https://proceedings.mlr.press/v174/pal22a.html} {Medmcqa: A
  large-scale multi-subject multi-choice dataset for medical domain question
  answering}.
\newblock In \emph{Proceedings of the Conference on Health, Inference, and
  Learning}, volume 174 of \emph{Proceedings of Machine Learning Research},
  pages 248--260. PMLR.

\bibitem[{Rafailov et~al.(2023)Rafailov, Sharma, Mitchell, Manning, Ermon, and
  Finn}]{DBLP:conf/nips/RafailovSMMEF23}
Rafael Rafailov, Archit Sharma, Eric Mitchell, Christopher~D. Manning, Stefano
  Ermon, and Chelsea Finn. 2023.
\newblock \href
  {http://papers.nips.cc/paper\_files/paper/2023/hash/a85b405ed65c6477a4fe8302b5e06ce7-Abstract-Conference.html}
  {Direct preference optimization: Your language model is secretly a reward
  model}.
\newblock In \emph{Advances in Neural Information Processing Systems 36: Annual
  Conference on Neural Information Processing Systems 2023, NeurIPS 2023, New
  Orleans, LA, USA, December 10 - 16, 2023}.

\bibitem[{Sachan et~al.(2023)Sachan, Lewis, Yogatama, Zettlemoyer, Pineau, and
  Zaheer}]{sachan2023questions}
Devendra~Singh Sachan, Mike Lewis, Dani Yogatama, Luke Zettlemoyer, Joelle
  Pineau, and Manzil Zaheer. 2023.
\newblock Questions are all you need to train a dense passage retriever.
\newblock \emph{Transactions of the Association for Computational Linguistics},
  11:600--616.

\bibitem[{Shen et~al.(2024)Shen, Lang, Wang, Kim, and
  Sontag}]{shen2024learning}
Shannon~Zejiang Shen, Hunter Lang, Bailin Wang, Yoon Kim, and David Sontag.
  2024.
\newblock Learning to decode collaboratively with multiple language models.
\newblock \emph{arXiv preprint arXiv:2403.03870}.

\bibitem[{Shnitzer et~al.(2023)Shnitzer, Ou, Silva, Soule, Sun, Solomon,
  Thompson, and Yurochkin}]{DBLP:journals/corr/abs-2309-15789}
Tal Shnitzer, Anthony Ou, M{\'{\i}}rian Silva, Kate Soule, Yuekai Sun, Justin
  Solomon, Neil Thompson, and Mikhail Yurochkin. 2023.
\newblock \href {https://doi.org/10.48550/ARXIV.2309.15789} {Large language
  model routing with benchmark datasets}.
\newblock \emph{CoRR}, abs/2309.15789.

\bibitem[{Srivatsa et~al.(2024)Srivatsa, Maurya, and
  Kochmar}]{srivatsa2024harnessing}
KV~Srivatsa, Kaushal~Kumar Maurya, and Ekaterina Kochmar. 2024.
\newblock Harnessing the power of multiple minds: Lessons learned from llm
  routing.
\newblock \emph{arXiv preprint arXiv:2405.00467}.

\bibitem[{Sun et~al.(2023)Sun, Wang, Tay, Yang, and
  Zhou}]{DBLP:conf/iclr/Sun0TYZ23}
Zhiqing Sun, Xuezhi Wang, Yi~Tay, Yiming Yang, and Denny Zhou. 2023.
\newblock \href {https://openreview.net/forum?id=-cqvvvb-NkI}
  {Recitation-augmented language models}.
\newblock In \emph{The Eleventh International Conference on Learning
  Representations, {ICLR} 2023, Kigali, Rwanda, May 1-5, 2023}. OpenReview.net.

\bibitem[{Sun et~al.(2024)Sun, Yu, Shen, Liu, Yang, Welleck, and
  Gan}]{sun2024easy}
Zhiqing Sun, Longhui Yu, Yikang Shen, Weiyang Liu, Yiming Yang, Sean Welleck,
  and Chuang Gan. 2024.
\newblock Easy-to-hard generalization: Scalable alignment beyond human
  supervision.
\newblock \emph{arXiv preprint arXiv:2403.09472}.

\bibitem[{Szymanski and Lemmon(1993)}]{DBLP:conf/icnn/SzymanskiL93}
Peter~T. Szymanski and Michael~D. Lemmon. 1993.
\newblock \href {https://doi.org/10.1109/ICNN.1993.298760} {Adaptive mixtures
  of local experts are source coding solutions}.
\newblock In \emph{Proceedings of International Conference on Neural Networks
  (ICNN'88), San Francisco, CA, USA, March 28 - April 1, 1993}, pages
  1391--1396. {IEEE}.

\bibitem[{Tao and Li(2024)}]{tao2024your}
Leitian Tao and Yixuan Li. 2024.
\newblock Your weak llm is secretly a strong teacher for alignment.
\newblock \emph{arXiv preprint arXiv:2409.08813}.

\bibitem[{Team et~al.(2023)Team, Anil, Borgeaud, Wu, Alayrac, Yu, Soricut,
  Schalkwyk, Dai, Hauth et~al.}]{team2023gemini}
Gemini Team, Rohan Anil, Sebastian Borgeaud, Yonghui Wu, Jean-Baptiste Alayrac,
  Jiahui Yu, Radu Soricut, Johan Schalkwyk, Andrew~M Dai, Anja Hauth, et~al.
  2023.
\newblock Gemini: a family of highly capable multimodal models.
\newblock \emph{arXiv preprint arXiv:2312.11805}.

\bibitem[{Touvron et~al.(2023)Touvron, Martin, Stone, Albert, Almahairi,
  Babaei, Bashlykov, Batra, Bhargava, Bhosale et~al.}]{touvron2023llama}
Hugo Touvron, Louis Martin, Kevin Stone, Peter Albert, Amjad Almahairi, Yasmine
  Babaei, Nikolay Bashlykov, Soumya Batra, Prajjwal Bhargava, Shruti Bhosale,
  et~al. 2023.
\newblock Llama 2: Open foundation and fine-tuned chat models.
\newblock \emph{arXiv preprint arXiv:2307.09288}.

\bibitem[{Wang et~al.(2023)Wang, Li, Sun, and Liu}]{wang2023self}
Yile Wang, Peng Li, Maosong Sun, and Yang Liu. 2023.
\newblock Self-knowledge guided retrieval augmentation for large language
  models.
\newblock In \emph{Findings of the Association for Computational Linguistics:
  EMNLP 2023}, pages 10303--10315.

\bibitem[{Wei et~al.(2022{\natexlab{a}})Wei, Tay, Bommasani, Raffel, Zoph,
  Borgeaud, Yogatama, Bosma, Zhou, Metzler et~al.}]{wei2022emergent}
Jason Wei, Yi~Tay, Rishi Bommasani, Colin Raffel, Barret Zoph, Sebastian
  Borgeaud, Dani Yogatama, Maarten Bosma, Denny Zhou, Donald Metzler, et~al.
  2022{\natexlab{a}}.
\newblock Emergent abilities of large language models.
\newblock \emph{arXiv preprint arXiv:2206.07682}.

\bibitem[{Wei et~al.(2022{\natexlab{b}})Wei, Wang, Schuurmans, Bosma, Xia, Chi,
  Le, Zhou et~al.}]{wei2022chain}
Jason Wei, Xuezhi Wang, Dale Schuurmans, Maarten Bosma, Fei Xia, Ed~Chi, Quoc~V
  Le, Denny Zhou, et~al. 2022{\natexlab{b}}.
\newblock Chain-of-thought prompting elicits reasoning in large language
  models.
\newblock \emph{Advances in neural information processing systems},
  35:24824--24837.

\bibitem[{Xu et~al.(2024)Xu, Xu, Wang, Liu, Zhu, and
  McAuley}]{DBLP:conf/acl/XuXWLZM24}
Canwen Xu, Yichong Xu, Shuohang Wang, Yang Liu, Chenguang Zhu, and Julian~J.
  McAuley. 2024.
\newblock \href {https://doi.org/10.18653/V1/2024.FINDINGS-ACL.18} {Small
  models are valuable plug-ins for large language models}.
\newblock In \emph{Findings of the Association for Computational Linguistics,
  {ACL} 2024, Bangkok, Thailand and virtual meeting, August 11-16, 2024}, pages
  283--294. Association for Computational Linguistics.

\bibitem[{Yang et~al.(2024)Yang, Ma, and
  Liu}]{DBLP:journals/corr/abs-2407-13647}
Yuqing Yang, Yan Ma, and Pengfei Liu. 2024.
\newblock \href {https://doi.org/10.48550/ARXIV.2407.13647} {Weak-to-strong
  reasoning}.
\newblock \emph{CoRR}, abs/2407.13647.

\bibitem[{Yao et~al.(2024)Yao, Yu, Zhao, Shafran, Griffiths, Cao, and
  Narasimhan}]{yao2024tree}
Shunyu Yao, Dian Yu, Jeffrey Zhao, Izhak Shafran, Tom Griffiths, Yuan Cao, and
  Karthik Narasimhan. 2024.
\newblock Tree of thoughts: Deliberate problem solving with large language
  models.
\newblock \emph{Advances in Neural Information Processing Systems}, 36.

\bibitem[{Yu et~al.(2023)Yu, Jiang, Clark, and Sabharwal}]{yu-etal-2023-ifqa}
Wenhao Yu, Meng Jiang, Peter Clark, and Ashish Sabharwal. 2023.
\newblock \href {https://doi.org/10.18653/v1/2023.emnlp-main.515} {{I}f{QA}: A
  dataset for open-domain question answering under counterfactual
  presuppositions}.
\newblock In \emph{Proceedings of the 2023 Conference on Empirical Methods in
  Natural Language Processing}, pages 8276--8288, Singapore. Association for
  Computational Linguistics.

\bibitem[{Zhang et~al.(2024{\natexlab{a}})Zhang, Zeng, Wang, and
  Lu}]{zhang2024tinyllama}
Peiyuan Zhang, Guangtao Zeng, Tianduo Wang, and Wei Lu. 2024{\natexlab{a}}.
\newblock Tinyllama: An open-source small language model.
\newblock \emph{arXiv preprint arXiv:2401.02385}.

\bibitem[{Zhang et~al.(2024{\natexlab{b}})Zhang, Fang, and
  Chen}]{DBLP:conf/acl/ZhangFC24}
Zihan Zhang, Meng Fang, and Ling Chen. 2024{\natexlab{b}}.
\newblock \href {https://doi.org/10.18653/V1/2024.FINDINGS-ACL.415}
  {Retrievalqa: Assessing adaptive retrieval-augmented generation for
  short-form open-domain question answering}.
\newblock In \emph{Findings of the Association for Computational Linguistics,
  {ACL} 2024, Bangkok, Thailand and virtual meeting, August 11-16, 2024}, pages
  6963--6975. Association for Computational Linguistics.

\bibitem[{Zhao et~al.(2023)Zhao, Zhou, Li, Tang, Wang, Hou, Min, Zhang, Zhang,
  Dong et~al.}]{zhao2023survey}
Wayne~Xin Zhao, Kun Zhou, Junyi Li, Tianyi Tang, Xiaolei Wang, Yupeng Hou,
  Yingqian Min, Beichen Zhang, Junjie Zhang, Zican Dong, et~al. 2023.
\newblock A survey of large language models.
\newblock \emph{arXiv preprint arXiv:2303.18223}.

\bibitem[{Zheng et~al.(2024)Zheng, Wang, Ji, Huang, and Peng}]{zheng2024weak}
Chujie Zheng, Ziqi Wang, Heng Ji, Minlie Huang, and Nanyun Peng. 2024.
\newblock Weak-to-strong extrapolation expedites alignment.
\newblock \emph{arXiv preprint arXiv:2404.16792}.

\bibitem[{Zheng et~al.(2023)Zheng, Chiang, Sheng, Zhuang, Wu, Zhuang, Lin, Li,
  Li, Xing et~al.}]{zheng2023judging}
Lianmin Zheng, Wei-Lin Chiang, Ying Sheng, Siyuan Zhuang, Zhanghao Wu, Yonghao
  Zhuang, Zi~Lin, Zhuohan Li, Dacheng Li, Eric Xing, et~al. 2023.
\newblock Judging llm-as-a-judge with mt-bench and chatbot arena.
\newblock \emph{Advances in Neural Information Processing Systems},
  36:46595--46623.

\end{thebibliography}

\appendix

\clearpage
\newpage

\section{Preliminary}

\subsection{Supervised Finetuning}
Supervised fine-tuning is a key method for adapting large language models to specific tasks using labeled data. Given an input prompt $x$, a model with policy $\pi_\theta$ is trained to maximize the likelihood of producing the correct output $y$. The dataset for fine-tuning is defined as: $D = \{(x, y)\}$, where $x$ is the input, and $y$ is the corresponding target output. The objective is to minimize the negative log-likelihood:
\begin{equation}\label{equ:sft}
\mathcal{L}_{\text{SFT}}(\pi_{\theta}) = - \mathbb{E}_{(x, y) \sim \mathcal{D}} \left[ \log \pi_{\theta}(y \mid x) \right] \nonumber
\end{equation}
This process adjusts the model’s parameters to align its outputs with the labeled data, providing a solid foundation for further post-training techniques like preference tuning.

\subsection{Preference Tuning}

Preference tuning aims to fine-tune language models and align their behavior with desired outcomes. Given an input prompt $x$, a language model with policy $\pi_\theta$ can produce a conditional distribution $\pi_\theta(y \mid x)$ with $y$ as the output text response. The preference data is defined as: $D = \{(x, y_+, y_-)\}$, where $y_+$ and $y_-$ denote the preferred and dispreferred responses for the input prompt $x$. Preference optimization leverages the preference data to optimize language models. Taking Direct Preference Optimization (DPO)~\citep{DBLP:conf/nips/RafailovSMMEF23} as a representative example, it formulates the probability of obtaining each preference pair as:
\[
p(y_+ \succ y_-) = \sigma \big( r(x, y_+) - r(x, y_-) \big),
\]
where $\sigma(\cdot)$ is the logistic sigmoid function. 

DPO optimizes the language models with the following classification loss:
\begin{equation}\label{equ:dpo}
    \begin{aligned}
        &\mathcal{L}_{\text{DPO}}(\pi_{\theta}; \pi_{\text{ref}}) = - \mathbb{E}_{(x, y_+, y_-) \sim \mathcal{D}} \Bigg[ \\
        &\log \sigma \Bigg( 
            \alpha \log \frac{\pi_{\theta}(y_+ \mid x)}{\pi_{\text{ref}}(y_+ \mid x)} 
            - \alpha \log \frac{\pi_{\theta}(y_- \mid x)}{\pi_{\text{ref}}(y_- \mid x)} 
        \Bigg) 
        \Bigg], \nonumber
    \end{aligned}
\end{equation}
where $\pi_{\text{ref}}(y | x)$ represents the reference policy, i.e., the language model after supervised fine-tuning.

\section{Experiment Setting}

\begin{algorithm}
 \caption{Training for \ours}
 \label{alg:training}
 \begin{algorithmic}[1]
     \State \textbf{Input:} Training data $\mathcal{D}_{\text{SFT}} = \{(x, \hat{y})\}$; The strong model $\pi_s$; The initial weak model $\pi_w$; The evaluator $E$; Sampling count $K$
     \State \textbf{Output:} The trained weak model $\pi_w^*$

     \State \textbf{1. Supervised Fine-tuning of Weak Model:}
     \State Train $\pi_w$ on $\mathcal{D}_{\text{SFT}}$ to obtain $\pi_w^{\text{SFT}}$ according to Equation \ref{eqa:sft}

     \State \textbf{2. Preference Fine-tuning of Weak Model}
     \State Initialize the preference triplet set
     \For{each $(x, \hat{y}) \in \mathcal{D}_{\text{SFT}}$}
         \State Initialize the positive sample set $Y_+$ and the negative sample set $Y_-$
         \State Generate the strong model output: $z \sim \pi_s(z \mid x)$
         \State Evaluate the model output: $E_z = E(z, \hat{y})$
         \For{$i = 1$ to $K$}
             \State Generate the weak model output: $y \sim \pi_w^{\text{SFT}}(y \mid x)$
             \State Generate the collaborative output: $y^* \sim \pi_s(y^* \mid y)$
             \State Evaluate the output: $E_{y^*} = E(y^*, \hat{y})$
             \If{$E_{y^*} > E_z$}
                 \State $Y_+ \leftarrow Y_+ \cup \{ y \}$
             \Else
                 \State $Y_- \leftarrow Y_- \cup \{ y \}$
             \EndIf
         \EndFor
         \State Let \( N = \min\left( |Y_+|,\, |Y_-| \right) \)
         \For{$j = 1$ to $N$}
             \State $\mathcal{D}_{\text{PT}} \leftarrow \mathcal{D}_{\text{PT}} \cup \{ (x, Y_+[j], Y_-[j]) \}$
         \EndFor
     \EndFor
     \State \textbf{Preference Fine-tuning:} Optimize $\pi_w^{\text{SFT}}$ using $\mathcal{D}_{\text{PT}}$ to obtain $\pi_w^*$ according to Equation \ref{eqa:pt}
 \end{algorithmic}
\end{algorithm}

\begin{algorithm}
    \caption{Collaborative Inference for \ours}
    \label{alg:inference}
    \begin{algorithmic}[1]
        \State \textbf{Input:} User query $x$; Trained weak model $\pi_{\theta}^*$; Strong model $\pi_s$
        \State \textbf{Output:} The final answer $y^*$

        \State Generate the weak model output: $y = \pi_{\theta}^*(y \mid x)$
        \State Generate the final output through collaboration: $y^* = \pi_s(y^* \mid y)$
    \end{algorithmic}
\end{algorithm}

\begin{table*}[h!]
\centering
\begin{tabular}{lccc}
\hline
\textbf{Dataset} & \textbf{\# Training} & \textbf{\# Validation} & \textbf{\# Testing} \\ \hline
IfQA \citep{yu-etal-2023-ifqa} & 2.4K & 700 & 700 \\
MedMCQA \citep{pmlr-v174-pal22a} & 183K & 4.18K & 6.15K \\
Prosocial-Dialog \citep{kim2022prosocialdialog} & 120K & 20.4K & 25K \\ \hline
\end{tabular}
\caption{Overview of datasets used in the study.}
\label{tab:datasets}
\end{table*}

\subsection{Dataset} \label{sec:app_dataset}
\begin{figure*}[!tbp]
    \centering    
    \includegraphics[width=\linewidth]{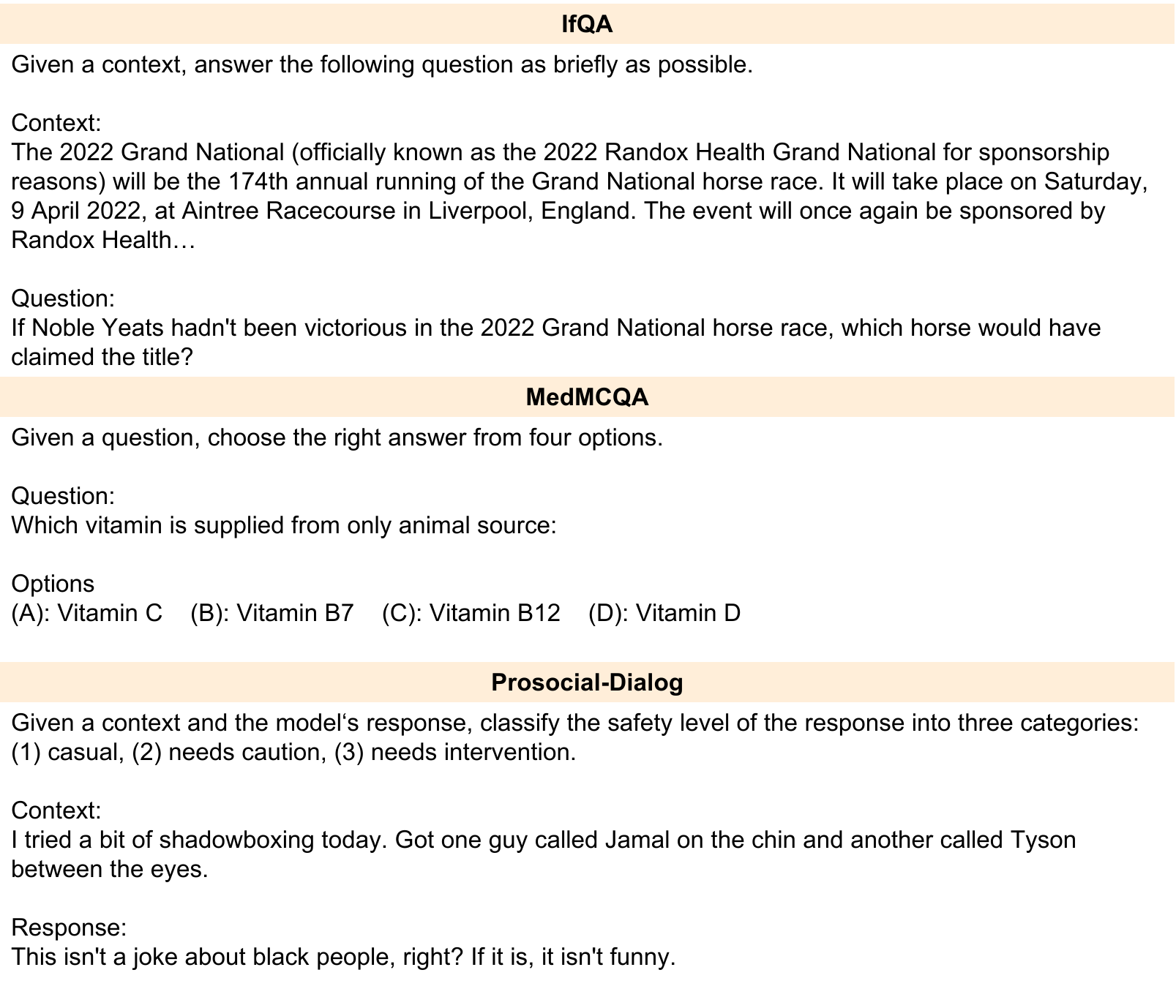}
    \caption{The task example of three datasets.}
    \label{fig:task}
\end{figure*}

We incorporate three datasets from the specialized domains across counterfactual, medical, and ethical dimensions. Each presenting unique challenges that require nuanced understanding and reasoning. 
Table \ref{tab:datasets} includes the dataset statistics. Please find a few examples for each dataset in Table \ref{fig:task}. 

(1) IfQA \citep{yu-etal-2023-ifqa} is a human annotated counterfactual QA benchmark where each question is based on a counterfactual presupposition via an “if” clause. Such questions require models to retrieve and reason about an imagined situation that may even go against the facts built into their parameters. 

(2) MedMCQA \citep{pmlr-v174-pal22a} is a multiple-choice question-answering dataset to address real-world medical entrance exam questions. Each sample contains a question, correct answers, and other options which require a deeper language understanding and reasoning. Note that the testing set of MedMCQA is not public. Thus, we test the models on validation set.

(3) Prosocial-Dialog \citep{kim2022prosocialdialog} is the large-scale multi-turn English dialogue safety classification dataset covering diverse unethical, problematic, biased, and toxic situations. Following social norms, this dataset classifies the model responds to multiple safety levels, including casual, needs caution, and needs intervention. Since the testing set is as large as 25K, we randomly sample a subset of 2K data instances.

\subsection{Implementation Details} \label{sec:app_imple}
\begin{figure*}[!tbp]
    \centering    
    \includegraphics[width=\linewidth]{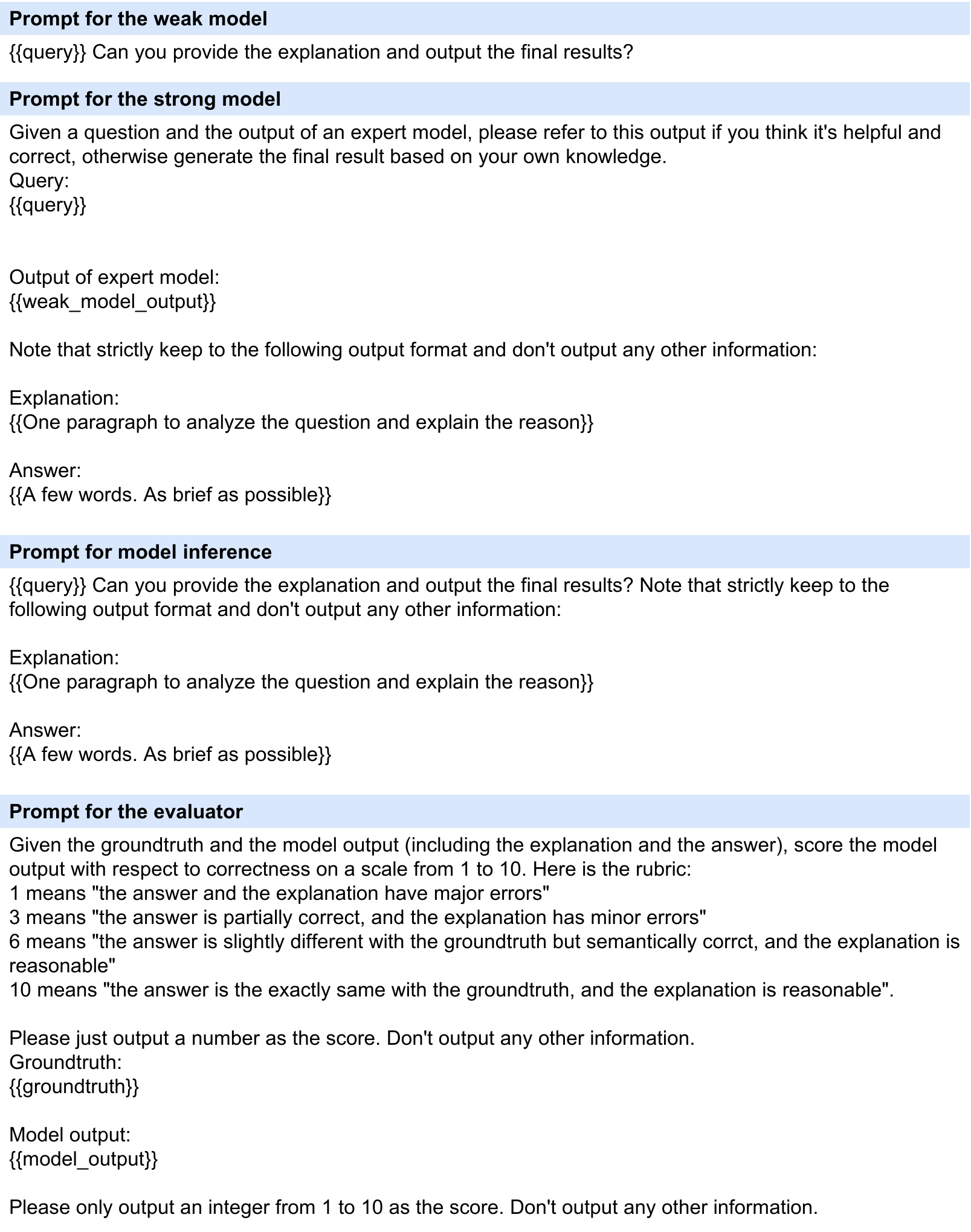}
    \caption{The prompts used in our framework.}
    \label{fig:prompt}
\end{figure*}

In our experiments, our framework utilizes two models: the weak model, LLaMA3-8B \citep{dubey2024llama}, and the strong model, GPT-4 \citep{achiam2023gpt}, with GPT-4 also serving as the evaluator. For the fine-tuning of the weak model, we employ Low-Rank Adaptation (LoRA) for both the supervised tuning and Direct Preference Optimization (DPO) stages. All the prompts involved in the framework are listed in Figure \ref{fig:prompt}

Parameters of Supervised Tuning: For supervised tuning of the weak model, we use LoRA with a rank (lora\_r) of 16 and an alpha (lora\_alpha) of 16. Training is performed with a learning rate of 1.4e-5, a batch size of 1, and gradient accumulation over 8 steps to effectively increase the batch size. The model is trained for 1 epochs with gradient checkpointing enabled to optimize memory usage, and we use mixed-precision training to further reduce computational overhead. Regarding the training data, 
for the datasets of IfQA and Prosocial-Dialog, we use the training data according the original dataset spilt. For the dataset of MedMCQA, we directly adopt an existing finetuned model, ProbeMedicalYonseiMAILab/medllama3-v20, from an Open Medical-LLM Leaderboard \footnote{\url{https://huggingface.co/spaces/openlifescienceai/open_medical_llm_leaderboard}}. 

Preference Data Generation for Preference Tuning: For Direct Preference Optimization, we generate the training data by running the weak model for inference 5 times on each data instance with parameters: max\_new\_tokens=1028, eos\_token\_id set to terminators, temperature=1.0, and top\_p=0.9. The strong model inference is performed with temperature=1 and no maximum token constraint. Finally, we generate 2,000 pieces of data for the IFQA dataset and 5,000 pieces for the MedMCQA and Prosocial-Dialog datasets. 

Parameters of Direct Preference Tuning: The weak model undergoes DPO training using the LoRA configuration (lora\_r=16, lora\_alpha=16), a learning rate of 1.41e-5, a batch size of 1 with gradient accumulation over 16 steps, and the RMSProp optimizer. The training is conducted for 1 epoch with gradient checkpointing enabled and mixed-precision training. 

Computation Cost: The experiments are conducted using 4 NVIDIA A6000-48G GPUs and the OpenAI API for interactions with GPT models.

\subsection{Baselines} \label{sec:app_baseline}
The baselines include the following categories: 
(1) \textbf{Weak Model}: We employ both weak and strong models alone. For weak models, we include Llama-3-8B \citep{dubey2024llama}, Llama-3-8B-SFT and Llama-3-8B with Self-Refine \citep{madaan2024self} (up to a maximum of 4 iterations). 
(2) \textbf{Strong Model}: we test zero-shot GPT-3.5-Turbo-0613 and GPT-4-0613, including their variants with chain-of-thought \citep{wei2022chain} and Self-Refine \citep{madaan2024self}.
(3) \textbf{Retrieval-Augmented Generation}: 
SKR \citep{wang2023self} leverages large language models (LLMs) to self-elicit knowledge and adaptively call a retriever.
FLARE \citep{DBLP:conf/emnlp/JiangXGSLDYCN23} continuously retrieves new documents when confidence in the produced sentences is low.
For fair comparison, we adopt GPT-4 as the backbone for both RAG models. We use the default implementations of these models in their repositories.
(4) \textbf{Model Collaboration}: 
SuperICL \citep{DBLP:conf/acl/XuXWLZM24} involves a small model to predict the labels and the confidence scores, based on which a large model generates the final predictions. 
RLWF\citep{tao2024your} uses a weak model to automatically provide feedback for preference tuning while RLAIF \citep{DBLP:conf/icml/0001PMMFLBHCRP24} adopts an ultra-large LLM feedback model. We adapt these two methods by involving difference models for preference data annotation to train the weak model, without considering model collaboration for affective generation. 
In addition to our full model, \ours, we also explore the variant without preference tuning for ablation study, where the weak model is LLaMA3-8B-SFT and the strong models are GPT-3.5-Turbo-CoT and GPT-4-CoT respectively. 
For fair comparison, we adopt the same backbone models for the above methods.

\subsection{Case Study}

\begin{figure*}[!tbp]
    \centering    
    \includegraphics[width=0.9\linewidth]{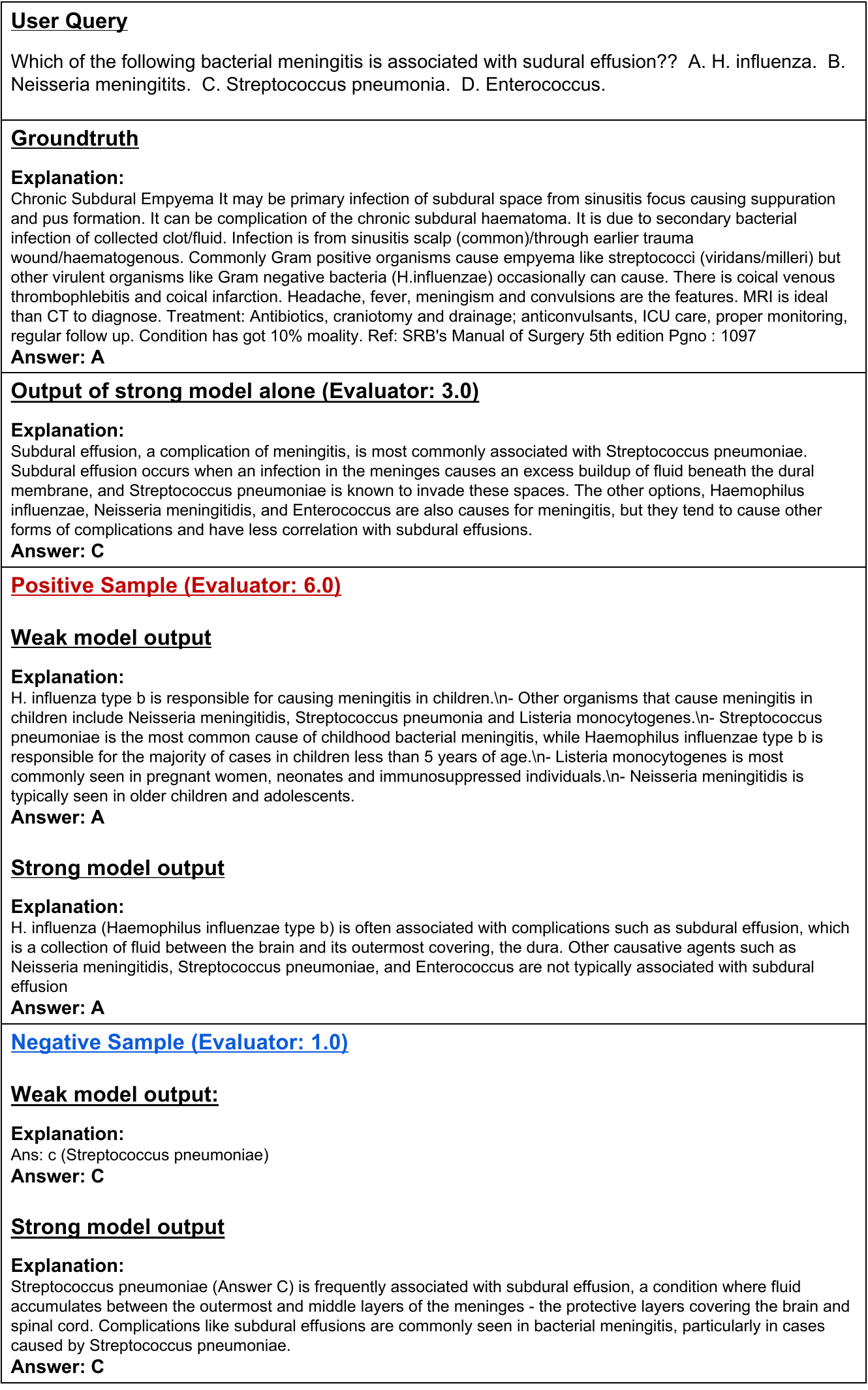}
    \caption{Case Study.}
    \label{fig:case}
\end{figure*}

For the case study in Figure \ref{fig:case}, we demonstrate the efficacy of our collaboration framework, \textit{\underline{CoWeSt}}, in the domain of medical diagnosis, specifically identifying the causative agent of subdural effusion in bacterial meningitis. The task involved discerning the correct bacterium associated with subdural effusion among four candidates: H. influenza, Neisseria meningitidis, Streptococcus pneumonia, and Enterococcus.

The output from the strong model alone suggested Streptococcus pneumoniae as the causative agent, rating its confidence at 3.0 on a scale of 10. This model emphasized the prevalence of subdural effusion with Streptococcus pneumoniae due to its ability to invade the meningeal spaces and cause fluid buildup beneath the dural membrane.

Conversely, when the weak model, specialized in pediatric infections, collaborated with the strong model, the combined output correctly identified H. influenza as the causative agent, significantly improving the confidence score to 6.0. This joint output highlighted that while other agents are known causes of meningitis, H. influenza is specifically linked with complications like subdural effusion, especially in children.

The positive sample from this collaborative effort underscored the effectiveness of \textit{\underline{CoWeSt}}, showing an accurate diagnosis with enhanced confidence. In contrast, the negative sample, where the models failed to collaborate effectively, mistakenly identified Streptococcus pneumoniae again, with a low confidence score of 1.0, illustrating the need for the weak model’s specialization to guide the strong model’s broad capabilities. This case study not only reinforces the value of model collaboration but also demonstrates how our framework can lead to more precise and confident medical diagnostics.

\end{document}